\documentclass{article}

\usepackage{PRIMEarxiv}

\usepackage[utf8]{inputenc} % allow utf-8 input
\usepackage[T1]{fontenc}    % use 8-bit T1 fonts
\usepackage{hyperref}       % hyperlinks
\usepackage{url}            % simple URL typesetting
\usepackage{booktabs}       % professional-quality tables
\usepackage{amsfonts}       % blackboard math symbols
\usepackage{nicefrac}       % compact symbols for 1/2, etc.
\usepackage{microtype}      % microtypography
\usepackage{lipsum}
\usepackage{fancyhdr}       % header
\usepackage{graphicx}       % graphics
\usepackage{amsmath} 
\usepackage{indentfirst} 
\usepackage{amssymb}

\graphicspath{{media/}}     % organize your images and other figures under media/ folder

\title{Enhancing Cognitive Diagnosis by Modeling Learner Cognitive Structure State}

\author{
  Zhifu Chen \\
  Northeast Normal University \\
  \texttt{zhifuchen@nenu.edu.cn} \\
   \And
     Hengnian Gu \\
  Northeast Normal University \\
  \texttt{guhn546@nenu.edu.cn} \\
   \And
     Jin Peng Zhou \\
  Cornell University \\
  \texttt{jpzhou@cs.cornell.edu} \\
   \And
  Dongdai Zhou\thanks{Corresponding author.} \\
  Northeast Normal University \\
  \texttt{ddzhou@nenu.edu.cn} \\
}

\begin{document}
\maketitle

\begin{abstract}
Cognitive diagnosis represents a fundamental research area within intelligent education, with the objective of measuring the cognitive status of individuals.Theoretically, an individual's cognitive state is essentially equivalent to their cognitive structure state. Cognitive structure state comprises two key components: \textbf{knowledge state(KS) and knowledge structure state(KUS)}.The knowledge state reflects the learner's mastery of individual concepts, a widely studied focus within cognitive diagnosis. In contrast, the knowledge structure state—representing the learner's understanding of the relationships between concepts—remains inadequately modeled.A learner's cognitive structure is essential for promoting meaningful learning and shaping academic performance. Although various methods have been proposed, most focus on assessing KS and fail to assess KUS.  To bridge this gap, we propose an innovative and effective framework—\textbf{CSCD}(\underline{C}ognitive \underline{S}tructure State-based \underline{C}ognitive \underline{D}iagnosis)—which introduces a novel framework to modeling learners' cognitive structures in diagnostic assessments, thereby offering new insights into cognitive structure modeling. Specifically, we employ an edge-feature-based graph attention network to represent the learner's cognitive structure state, effectively integrating KS and KUS.Extensive experiments conducted on real datasets demonstrate the superior performance of this framework in terms of diagnostic accuracy and interpretability.
\end{abstract}

% keywords can be removed

\section{Introduction}
Cognitive diagnosis (CD) is a pivotal and foundational research domain, extensively applied in real-world scenarios such as gaming\cite{chen2016predicting}, medical diagnostics\cite{guo2017modeling}, and education, particularly in intelligent education systems\cite{anderson2014engaging}. Cognitive diagnosis represents a fundamental research area within intelligent education, with the objective of measuring the cognitive status of individuals. Cognitive diagnosis seeks to analyze learners' mastery of knowledge throughout the learning process\cite{liu2018fuzzy}.

\begin{figure}[h]
  \centering
  \includegraphics{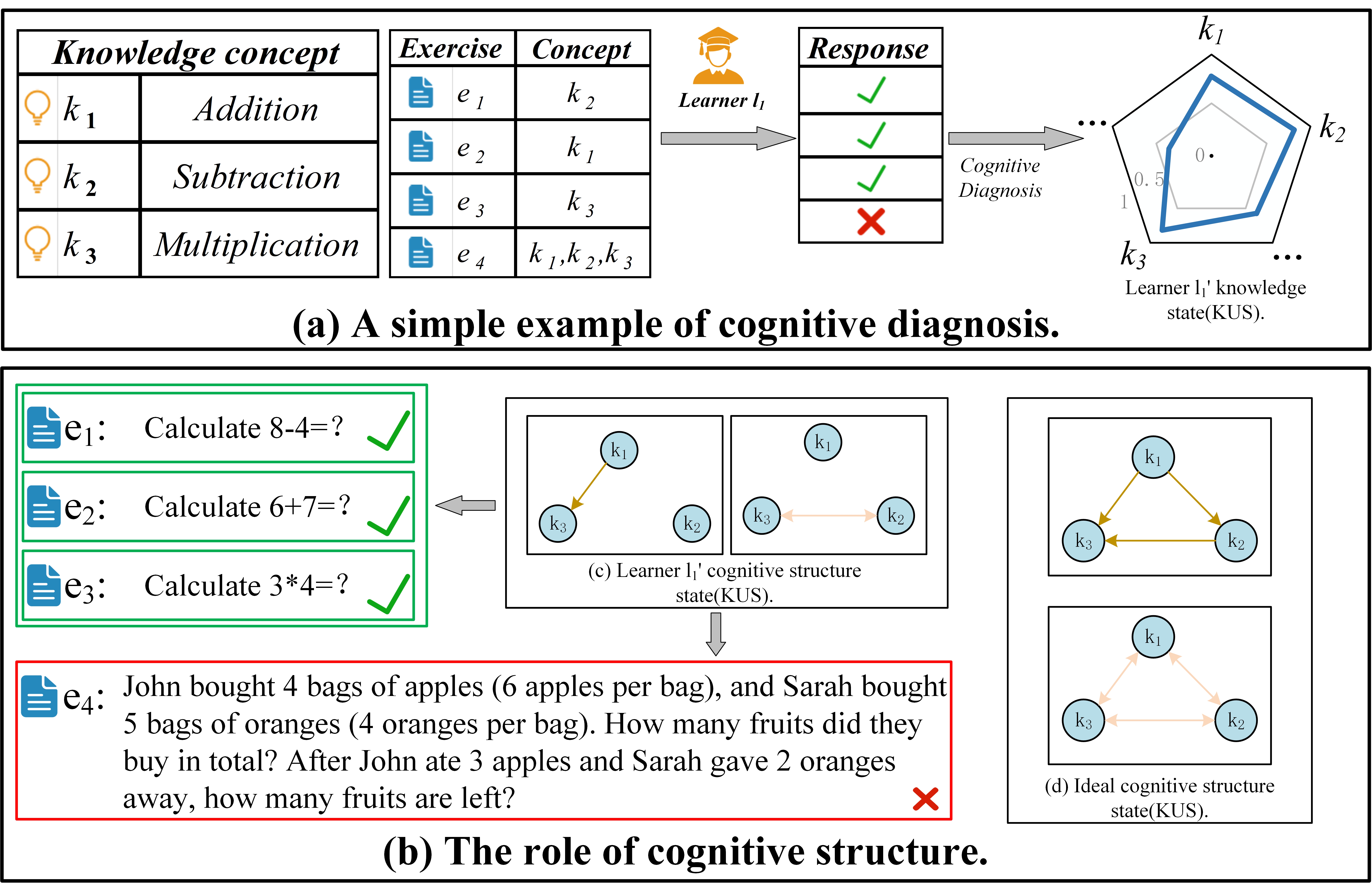}
  \caption{The illustration of (a) A simple example of cognitive diagnosis; (b) The importance of the knowledge structure state(KUS).}
  \label{exampleId1}
\end{figure}

Figure \ref{exampleId1}-(a) presents an example of cognitive diagnosis: learners generally begin by completing a series of exercises (e.g., e1 to e4) and recording their responses (correct or incorrect). Based on this data, the objective of the research is to infer the learners' actual knowledge state concerning related concepts. The results of cognitive diagnosis not only provide the foundation for personalized services (such as exercise recommendations and targeted training)\cite{kuh2011piecing}, but are also extensively utilized in the development and optimization of areas such as course recommendation, learner profiling, and computerized adaptive testing. Through quantitative evaluation, CD models can accurately predict learners' responses to test exercises and present their proficiency with various knowledge state in an intuitive format, such as radar charts.	

We now define the knowledge state and the knowledge structure state as follows:

\textbf{Knowledge State (KS):} The knowledge state (KS) refers to the learner's mastery of individual concepts.

\textbf{Knowledge Structure State (KUS):} The knowledge structure state (KUS) refers to the learner's understanding of the relationships between concepts, as well as their mastery of these interrelationships.

However, a notable issue is evident in Figure \ref{exampleId1}-(a): Exercise 4 incorporates the knowledge concepts from Exercises 1, 2, and 3. If a learner answers Exercises 1, 2, and 3 correctly, they should also be able to answer Exercise 4 accurately. However, the learner exhibits completely contrary behavior when answering Exercise 4. In previous studies, this situation is often attributed merely to the problem's difficulty or other external factors. However, the underlying reason for learner \(l_1\)'s errors in answering exercise 4 lies in his inability to effectively integrate the knowledge concepts \(k_1\), \(k_2\), and \(k_3\), resulting in the lack of a cohesive knowledge structure. In other words, while the learner has mastered \( k_1 \), \( k_2 \), and \( k_3 \), they have not fully understood the relationships between these concepts, such as predecessor-successor relationships and dependencies. From the perspectives of cognitive psychology and educational psychology, the cognitive structure of learner \( l_1 \) is flawed.

Cognitive structure refers to the network of knowledge in a learner's mind, typically consisting of nodes (knowledge concepts) and edges (the relationships between these concepts). It is not only a central concept in cognitive psychology and educational psychology but also a fundamental element in both educational science and cognitive science. In Figure \ref{exampleId1}, the ideal cognitive structure for learner \( l_1 \) should resemble the one shown in Figure \ref{exampleId1}-(d). However, the actual cognitive structure state of learner \( l_1 \) during the learning process in this case is shown in Figure \ref{exampleId1}-(c). Compared to Figure \ref{exampleId1}-(d), the  learner \( l_1 \)'s cognitive structure is incomplete,which results in the incorrect response to \( e_4 \).

Since the mid-20th century, cognitive structure theory has been a cornerstone of educational theory and teaching practice. Evaluating learners' cognitive structure state allows for a more precise understanding of their cognitive strengths and weaknesses, thus providing strong support for personalized teaching. Ausubel et al.\cite{ausubel1978educational} emphasized that the most important factor influencing learning is the knowledge that learners have already acquired—that is, their cognitive structure. According to the SOLO taxonomy\cite{biggs2014evaluating}, based on cognitive structure theory, the learning process progresses from simplicity to complexity as learners' knowledge systems expand from isolated knowledge concepts to interconnected concepts, ultimately forming a complete cognitive structure. The SOLO taxonomy is widely applied in international learner assessment programs, such as PISA\cite{liu2024question}. Therefore, when measuring the cognitive status of individuals, both KS and KUS must be comprehensively assessed.

 However, cognitive psychology and educational psychology have yet to provide effective methods to quantitatively measure the cognitive structure state of individual learners. Although existing cognitive diagnosis methods have explored learners' cognitive states, most focus primarily on KS assessment and fail to adequately assess KUS. Choi et al. \cite{choi2020towards} introduced the concept of cognitive structure and defined it as the learner's knowledge level and the knowledge structure of learning items (e.g., prerequisite relationships), but their assessment still primarily focuses on KS. Gao et al. \cite{gao2021rcd} and Su et al. \cite{su2022graph} constructed a heterogeneous graph structure that leverages the dependency relationships between knowledge concepts to enhance the assessment of learners' KS. Li et al. \cite{li2022hiercdf} proposed the HierCDF framework, which models the impact of hierarchical knowledge structures on cognitive diagnosis. Song et al. \cite{song2023deep} focused on knowledge concept maps and the dependencies between knowledge concepts, while Jiao et al. \cite{jiao2023neural} revealed the concept dependencies within these maps. All these methods utilize the dependency relationships between knowledge concepts to assess learners' KS more precisely.

Essentially, these methods treat the objective and static relationships between knowledge concepts as features for diagnosing learners' progress in acquiring those concepts (or for enhancing their representations). However, their output ultimately focuses solely on assessing learners' mastery of individual knowledge concepts. While some researchers suggest that learners' mastery of the cognitive structure state, as an implicit feature, can be internalized into their mastery of knowledge state, these methods still fail to explicitly diagnose learners' mastery of both knowledge state and cognitive structure state. This limitation makes it difficult to provide accurate insights into learners' cognitive strengths and weaknesses. As a result, the precise identification of defects in learners' cognitive structures is hindered, making it challenging to deliver targeted learning recommendations or interventions. Therefore, we have revisited relevant theories, further explored the importance of revealing learners' cognitive structure state, and explicitly modeled both knowledge state and cognitive structure state in cognitive diagnosis.

Based on this, we propose the CSCD framework, which utilizes an edge-feature-based graph attention network to dynamically model and update both the knowledge state and the knowledge structure state. This approach effectively captures learners' cognitive structure state and applies it to diagnose their performance in answering questions. To the best of our knowledge, this is the first framework to systematically explore learners' cognitive structure state in the field of cognitive diagnosis. Extensive experiments on real-world datasets demonstrate that the framework offers significant advantages in terms of diagnostic accuracy and interpretability.

\section{Related Work}
\subsection{Cognitive diagnosis}
\textbf{Traditional cognitive diagnosis.} A significant body of literature has been devoted to cognitive diagnosis, including deterministic input, noise, and gate models (DINA)\cite{de2009dina}, item response theory (IRT)\cite{embretson2013item}, multidimensional IRT (MIRT)\cite{ackerman2014multidimensional}, and matrix factorization (MF)\cite{koren2009matrix}. Despite demonstrating some effectiveness, these approaches rely on manually defined interaction functions, which combine features of learners and exercises through multiplicative terms, such as logical functions\cite{embretson2013item} or inner products\cite{koren2009matrix}. However, this may be insufficient to capture the complex relationships between learners and exercises\cite{devellis2006classical}.

\textbf{Neural Cognitive Diagnosis.} In recent years, most neural cognitive diagnosis models employ neural networks to extract latent features of learners and exercise characteristics from data\cite{cheng2019dirt}, or to automatically learn higher-order nonlinear functions\cite{wang2020neural}. These models aim to capture various diagnostic features, such as contextual software features (e.g., family, school) modeled by ECD\cite{zhou2021modeling} and emotional states (e.g., boredom, confusion) modeled by ACD\cite{wang2024unified}. Other models focus on assessing and optimizing cognitive diagnosis tasks, such as ICD\cite{tong2022incremental} based on incremental learning, SCD\cite{shen2024symbolic} addressing the long-tail problem, ReliCD\cite{zhang2023relicd} alleviating data sparsity and noise, DCD\cite{chen2024disentangling} addressing the limitations of a labeled q-matrix, and UCD\cite{wang2024unified} for uncertainty assessment.

\textbf{Cognitive structure in cognitive diagnosis.} Numerous studies have introduced education-prior-based relational graphs, including Knowledge Concept (KC) graphs and item-concept association graphs, to improve the representation of learners and items. Gao et al.\cite{gao2021rcd} and Su et al.\cite{su2022graph} modeled heterogeneous graph structures for learning item knowledge, thoroughly exploring higher-order interactions between nodes and the dependency relationships between knowledge concepts in concept maps, thereby improving the representation of learners' cognitive states and item characteristics. Li et al.\cite{li2022hiercdf} proposed the HierCDF framework to model the influence of hierarchical knowledge structures on cognitive diagnosis. Song et al.\cite{song2023deep} focused on the effective integration of knowledge concept maps, concept dependencies, and item features. Jiao et al.\cite{jiao2023neural} revealed the relationships between knowledge concepts and items, as well as the concept dependencies within the knowledge concept map, thereby improving the representation of items and learner characteristics.Essentially, these methods treat the objective and static relationships between knowledge concepts as features for diagnosing learners' progress in acquiring those concepts (or for enhancing their representations). However, their output ultimately focuses solely on assessing learners' mastery of individual knowledge concepts. While some researchers suggest that learners' mastery of the cognitive structure state, as an implicit feature, can be internalized into their mastery of knowledge state, these methods still fail to explicitly diagnose learners' mastery of both knowledge state and cognitive structure state. This limitation makes it difficult to provide accurate insights into learners' cognitive strengths and weaknesses. As a result, the precise identification of defects in learners' cognitive structures is hindered, making it challenging to deliver targeted learning recommendations or interventions. Therefore, we have revisited relevant theories, further explored the importance of revealing learners' cognitive structure state, and explicitly modeled both knowledge state and cognitive structure state in cognitive diagnosis.

\subsection{Graph Neural Networks}
Deep learning, as a mainstream method in the field of artificial intelligence, has made significant progress across various application areas due to its powerful feature learning capabilities. Common neural network models, such as Convolutional Neural Networks (CNN)\cite{gu2018recent}, Recurrent Neural Networks (RNN)\cite{medsker2001recurrent}, and Generative Adversarial Networks (GAN)\cite{goodfellow2020generative}, have achieved great success in processing Euclidean data such as images, text, and speech. However, many real-world problems, such as social networks and knowledge graphs, involve complex graph-structured data that cannot be simply represented by traditional Euclidean space models. Graph Neural Networks (GNN) have emerged as an effective deep learning framework for processing graph-structured data.

Graph Convolutional Networks (GCN) are an important branch of GNNs. GCN extends the convolution operation from traditional regular data to irregular graph data. Unlike traditional GCN models, which treat all nodes equally, the introduction of attention mechanisms into GNNs, forming Graph Attention Networks (GAT), allows for assigning different attention weights to neighboring nodes of a target node, thereby highlighting the contributions of more significant neighbors. Velickovic et al.\cite{velickovic2017graph} were the first to introduce attention mechanisms into GNNs, proposing the GAT model, which learns the weights of different neighboring nodes through a neural network. During information aggregation, it focuses only on the nodes that have a significant impact on the target node while ignoring those with a smaller effect. This method dynamically adjusts the importance of neighboring nodes, significantly enhancing the model's performance. The basic structure of GAT incorporates an attention layer, which learns weights during the aggregation of neighboring nodes to update the target node representation. However, in many practical applications, the edges in a graph often contain rich information, yet most existing models still fail to fully leverage this edge information, limiting the comprehensive representation and analysis of graph-structured data.

To address this, Wang et al.\cite{wang2021egat} proposed the Edge-Featured Graph Attention Network (EGAT), based on GAT, which efficiently learns both node and edge features, generating more comprehensive and accurate feature representations. This collaborative optimization mechanism not only effectively integrates the feature information of nodes and edges in the graph but also significantly improves the model's ability to handle complex graph data. Through the design of multilayer attention mechanisms, EGAT is capable of capturing multiscale feature information, providing strong support for graph data representation learning and downstream tasks.

\section{Cognitive Structure-based Cognitive Diagnosis}
\subsection{Task Overview}
Let \( \mathcal{S} = \left\{  s_{1}, s_{2}, \ldots , s_{N} \right\} \) represent the set of \( N \) learners, \( E = \left\{  e_{1}, e_{2}, \ldots , e_{M} \right\} \) represent the set of \( M \) exercises, and \( C = \left\{  c_{1}, c_{2}, \ldots , c_{K} \right\} \) represent the set of \( K \) knowledge concepts. \( \mathcal{R}_{non-c} = \left\{  r_{c_{i} \leftrightarrow c_{j}} \mid c_{i} \in \mathcal{C}, c_{j} \in \mathcal{C}, i \neq j \right\} \) denotes the set of dependency correlations between concepts \( c_i \) and \( c_j \), and \( \mathcal{R}_{c} = \left\{  r_{c_{i} \rightarrow c_{j}} \mid c_{i} \in \mathcal{C}, c_{j} \in \mathcal{C}, i \neq j \right\} \) represents the set of predecessor-successor correlations between concepts \( c_i \) and \( c_j \). Each learner is assumed to independently select certain exercises for practice. We record the response records of a particular learner as a set of triplets \( \left( s, e, r_{se} \right) \), where \( s \in \mathcal{S} \), \( e \in E \), and \( r_{se} \) represents the score obtained by learner \( s \) on exercise \( e \). The relationship between test items and knowledge concepts is represented by the \( \mathrm{Q} \)-matrix, \( Q = \left( {q}_{ij}\right)_{M \times K} \). Specifically, \( {q}_{ij} = 1 \) indicates that test item \( i \) contains the \( j \)-th knowledge concepts, and \( {q}_{ij} = 0 \) indicates that test item \( i \) does not contain the \( j \)-th knowledge concepts, where \( i \in \{1, 2, \dots, M \} \) and \( j \in \{1, 2, \dots, K \} \).
Let \( \mathcal{L} \) denote the set of response records. Then, we provide a clear formulation of the cognitive diagnosis:

\textbf{Given:} learners’ response records \( \mathcal{L} \), exercises, dependency correlations \( \mathcal{R}_{non-c} \), and predecessor-successor correlations \( \mathcal{R}_{c} \).

\textbf{Goal:} To diagnose learners' cognitive structure state(i.e. knowledge state(KS) and knowledge structure state(KUS)) by modeling their performance prediction process.

\subsection{CSCD Framework}
\begin{figure}[h]
  \centering
  \includegraphics[width=0.6\linewidth]{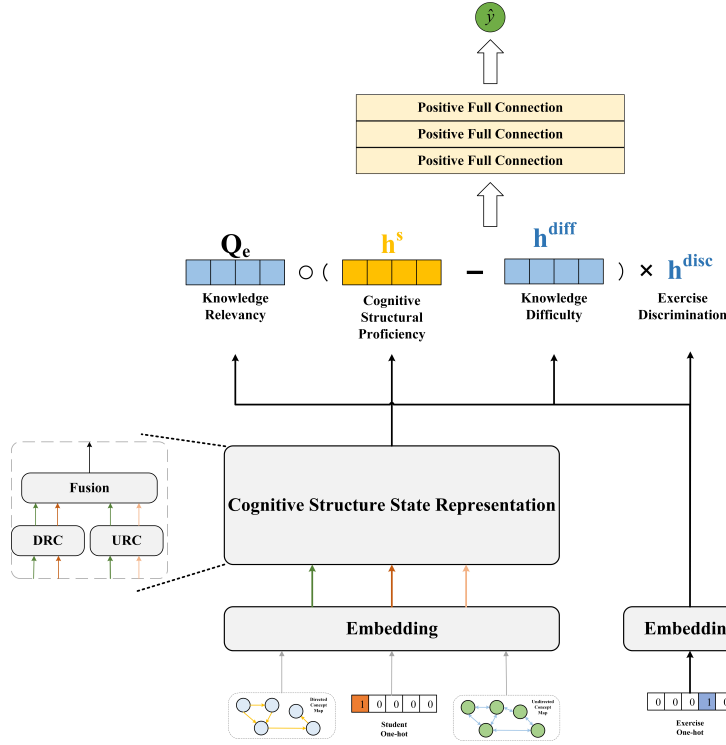}
  \caption{Structure of CSCD framework.}
  \label{exampleId2}
\end{figure}

The architecture of our proposed CSCD is illustrated in Figure \ref{exampleId2}.
\subsubsection{Embedding Module}
This paper employs one-hot encoding and embedding techniques to map knowledge concepts, predecessor-successor relationships among knowledge concepts, learners, and test items into a unified low-dimensional vector space, with final representations generated through the sigmoid function. The set of knowledge concepts is denoted as \( C = \left\{  {c}_{1},{c}_{2},\cdots ,{c}_{K}\right\} \), where \( K \) denotes the total number of knowledge concepts. First, each knowledge concepts \( {c}_{i} \) is encoded using one-hot encoding, as follows:  
\begin{equation}
{c}_{k} = \left\lbrack  {0,0,\cdots ,1,\cdots ,0}\right\rbrack   \in  {\mathbb{R}}^{K}
\end{equation}

where the \( i \)-th position is 1, and the rest are 0. Subsequently, \( {c}_{k} \) is embedded into a dense vector space to derive the embedding representation of knowledge concepts \( i \):
\begin{equation}
    {h}_{k} = \operatorname{sigmoid}\left( {{W}_{k} \cdot  {c}_{k} + {b}_{k}}\right) ,{h}_{k} \in  {\mathbb{R}}^{d}
\end{equation}

where \( {W}_{k} \in  {\mathbb{R}}^{d \times  K} \) is the weight matrix for the knowledge concepts embedding, \( {b}_{k} \in  {\mathbb{R}}^{d} \) is the bias vector, and \( d \) is the dimension of the embedding vector.Similarly, the embedding representations for predecessor-successor and dependency relationships are as follows:
\begin{equation}
h_{c_{i} \leftrightarrow c_{j} }= \operatorname{sigmoid}\left( {{W}_{r} \cdot  r_{c_{i} \leftrightarrow c_{j} }+ {b}_{r}}\right) ,h_{c_{i} \leftrightarrow c_{j} } \in  {\mathbb{R}}^{d}
\end{equation}
\begin{equation}
h_{c_{j} \rightarrow c_{i} }= \operatorname{sigmoid}\left( {{W}_{r} \cdot  r_{c_{j} \rightarrow c_{i} }+ {b}_{r}}\right) ,h_{c_{j} \rightarrow c_{i} } \in  {\mathbb{R}}^{d}
\end{equation}

The set of learners is denoted as \( S = \left\{  {s}_{1},{s}_{2},\cdots ,{s}_{N}\right\} \), where \( N \) represents the total number of learners. The one-hot encoding of the learners is expressed as:  
\begin{equation}
{ s}_{n} = \left\lbrack  {0,0,\cdots ,1,\cdots ,0}\right\rbrack   \in  {\mathbb{R}}^{N}
\end{equation}

The one-hot encoding of the learners is mapped into a low-dimensional dense space, yielding the learner embedding representation:
\begin{equation}
{h}_{n} = \operatorname{sigmoid}\left( {{W}_{s} \cdot  {s}_{n} + {b}_{s}}\right) ,{h}_{n} \in  {\mathbb{R}}^{d},
\end{equation}

where \( {W}_{s} \in  {\mathbb{R}}^{d \times  N} \) is the weight matrix for the learner embedding, \( {b}_{s} \in  {\mathbb{R}}^{d} \) is the bias vector, and \( d \) is the dimension of the embedding vector.

Based on the embedding representations of learners, knowledge concepts, and their relationships, the personalized embedding representation of each knowledge concepts for learner \( n \) is formulated as follows:  
For learner \( n \), the personalized embedding representation of each knowledge concepts is expressed as:  
\begin{equation}
{h}_{n,k} = \operatorname{sigmoid}\left( {{W}_{s,k} \cdot  \left\lbrack  {{h}_{n} \oplus  {h}_{k}}\right\rbrack   + {b}_{s,k}}\right)
\end{equation}

For learner \( n \), the relational embedding representation between knowledge concepts \( i \) and \( j \) is:
\begin{equation}
{h}_{n,{c_{i} \leftrightarrow c_{j} }} = \operatorname{sigmoid}\left( {{W}_{s,r} \cdot  \left\lbrack  {{h}_{n} \oplus  h_{c_{i} \leftrightarrow c_{j} }}\right\rbrack   + {b}_{s,r}}\right)
\end{equation}
\begin{equation}
{h}_{n,{c_{j} \rightarrow c_{i} }} = \operatorname{sigmoid}\left( {{W}_{s,r} \cdot  \left\lbrack  {{h}_{n} \oplus  h_{c_{j} \rightarrow c_{i} }}\right\rbrack   + {b}_{s,r}}\right)
\end{equation}

The set of test items is denoted as \( E = \left\{  {e}_{1},{e}_{2},\cdots ,{e}_{M}\right\} \), where \( M \) represents the total number of test items. The one-hot encoding of the test items is expressed as:
\begin{equation}
{e}_{m} = \left\lbrack  {0,0,\cdots ,1,\cdots ,0}\right\rbrack   \in  {\mathbb{R}}^{M}
\end{equation}
Based on the one-hot encoding of the test items, the difficulty embedding representation \( {h}^{\text{diff}} \) and discrimination embedding representation \( {h}^{\text{dics}} \) corresponding to each test item are derived using the embedding technique:
\begin{equation}
{h}^{\text{diff}} = \operatorname{sigmoid}\left( {{W}_{e} \cdot  {e}_{m} + {b}_{e}}\right)  \in  {\mathbb{R}}^{1 \times  K}
\end{equation}
\begin{equation}
{h}^{\text{dics}} = \operatorname{sigmoid}\left( {{W}_{e} \cdot  {e}_{m} + {b}_{e}}\right)  \in  {\mathbb{R}}^{1}
\end{equation}

For each exercise, the corresponding exercise factor is represented as \( Q_e \), derived directly from the pregiven Q-matrix:
\begin{equation}
Q_e = x^e \times \mathrm{Q}
\end{equation}

\subsubsection{Cognitive Structure State Representation Module}
In this section, we introduce the cognitive structure representation module, which consists of three components: DRC, URC, and the Fusion component. Both DRC and URC are implemented using EGAT, as EGAT can simultaneously learn the representations of knowledge state(KS) and knowledge structure state(KUS). Specifically, DRC learns the representations of knowledge state(KS) and predecessor-successor relationships of knowledge structure state(KUS), while URC learns the representations of knowledge state(KS) and dependency relationships of knowledge structure state(KUS).

\begin{figure}[h]
  \centering
  \includegraphics{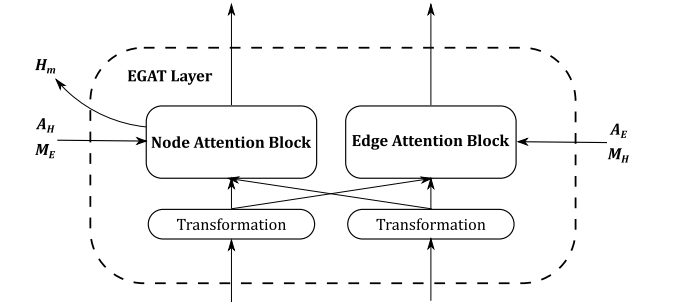}
  \caption{EGAT}
  \label{exampleId3}
\end{figure}

\textbf{EGAT}.We introduce the structure of EGAT, as shown in the figure \ref{exampleId3}. Each EGAT layer consists of a Node Attention Block and an Edge Attention Block. To illustrate, we will use predecessor-successor relationships as an example.The function representation of EGAT can be written as \( \mathcal{F}(\cdot) \).

In the Node Attention Block, the attention factor \( {a}_{{j} \rightarrow {i} } \) between node \( \mathrm{i} \) and node \( \mathrm{j} \) is computed as follows:

\begin{equation}
{a}_{{j} \rightarrow {i} } = \frac{\exp \left( {\operatorname{Leaky}\operatorname{Re}{LU}\left( {{\overrightarrow{\alpha }}^{\mathrm{T}}\left\lbrack  {{\overrightarrow{h}}_{i}\oplus {\overrightarrow{h}}_{j}\oplus {\overrightarrow{r}}_{{j} \rightarrow {i} }}\right\rbrack  }\right) }\right) }{\mathop{\sum }\limits_{{k \in  {N}_{i}}}\exp \left( {\operatorname{Leaky}\operatorname{Re}{LU}\left( {{\overrightarrow{\alpha }}^{\mathrm{T}}\left\lbrack  {{\overrightarrow{h}}_{i}\oplus {\overrightarrow{h}}_{k}\oplus {\overrightarrow{r}}_{{k} \rightarrow {i} }}\right\rbrack  }\right) }\right) }
\end{equation}

where \( {\overrightarrow{\alpha }}^{\mathrm{T}} \) is the weight vector parameter, \( {\overrightarrow{r}}_{{j} \rightarrow {i} } \) represents the edge feature vector between node \( \mathrm{i} \) and node \( \mathrm{j} \), \( \oplus \) denotes the vector concatenation operation, and LeakyReLU is a non-linear activation function. Based on the attention factor, the update rule for node features is:

\begin{equation}
{h}_{i}^{\prime } = \delta \left( {\mathop{\sum }\limits_{{j \in  {N}_{i}}}{a}_{{j} \rightarrow {i} }{h}_{j}}\right)
\end{equation}

In the Edge Attention Block, the attention factor \( {\beta }_{{q} \rightarrow {p} } \) between edge \( \mathrm{p} \) and its neighboring edge \( \mathrm{q} \) is computed as:

\begin{equation}
{\beta }_{{q} \rightarrow {p} } = \frac{\exp \left( {\operatorname{Leaky}\operatorname{Re}{LU}\left( {{\overrightarrow{b}}^{\mathrm{T}}\left\lbrack  {{\overrightarrow{r}}_{p}\oplus {\overrightarrow{r}}_{q}\oplus {\overrightarrow{h}}_{{q} \rightarrow {p} }}\right\rbrack  }\right) }\right) }{\mathop{\sum }\limits_{{k \in  {N}_{p}}}\exp \left( {\operatorname{Leaky}\operatorname{Re}{LU}\left( {{\overrightarrow{b}}^{\mathrm{T}}\left\lbrack  {{\overrightarrow{r}}_{r}\oplus {\overrightarrow{r}}_{k}\oplus {\overrightarrow{h}}_{{k} \rightarrow {p} }}\right\rbrack  }\right) }\right) }
\end{equation}

where \( {\overrightarrow{b}}^{\mathrm{T}} \) is the weight vector parameter, \( {\overrightarrow{h}}_{{q} \rightarrow {p} } \) represents the node feature vector between edge \( \mathrm{p} \) and its neighboring edge \( \mathrm{q} \), \( \oplus \) denotes the vector concatenation operation, and LeakyReLU is a non-linear activation function. Based on the attention factor, the update rule for edge features is:

\begin{equation}
{e}_{p}^{\prime } = \delta \left( {\mathop{\sum }\limits_{{q \in  {N}_{p}}}{\beta }_{{q} \rightarrow {p} }{h}_{q}}\right)
\end{equation}

The EGAT framework efficiently learns node and edge features through the parallel updates of the Node Attention Block and Edge Attention Block, yielding more comprehensive and precise feature representations. This collaborative optimization mechanism not only effectively integrates the feature information of nodes and edges in the graph but also substantially enhances the model's ability to handle complex graph data.

\textbf{Specializations for DRC and URC.} The knowledge state(KS) learned by learner \( \mathrm{n} \) and the dependency relationships of knowledge structure state(KUS) \( \mathrm{i} \) are updated as:
\begin{equation}
    {h}_{n,{\overleftrightarrow{k}}} ^{\prime },{h}_{n,{c_{i} \leftrightarrow c_{j} }}^{\prime } = \mathcal{F} \left({h}_{n,{\overleftrightarrow{k}}},{h}_{n,{c_{i} \leftrightarrow c_{j} }} \right)
\end{equation}

The knowledge state(KS) learned by learner \( \mathrm{n} \) and the predecessor-successor relationships of knowledge structure state(KUS)\( \mathrm{i} \) are updated as:
\begin{equation}
    {h}_{n,{\overrightarrow{k}}} ^{\prime },{h}_{n,{c_{j} \rightarrow c_{i} }}^{\prime } = \mathcal{F} \left({h}_{n,{\overrightarrow{k}}},{h}_{n,{c_{j} \rightarrow c_{i} }} \right)
\end{equation}

\textbf{Fusion.} This paper introduces a fusion operation that integrates the features of knowledge state(KS) and knowledge structure state(KUS), thereby generating the cognitive structure representation \( h^\mathrm{s} \) for learner \( n \) learning knowledge concepts \( i \). The specific fusion operation is mathematically defined as follows,taking the predecessor-successor relationship of knowledge structure state(KUS) as an example:

\begin{align}
    \begin{aligned}
\overrightarrow{h^{s}} &= \text{sigmoid}\left(\overrightarrow{W}\left[{h}_{n,{\overrightarrow{k}}} ^{\prime } \oplus h_{n,\overrightarrow{ip}}\right] + \overrightarrow{b}\right) \\
h_{n,\overrightarrow{ip}} &= \sum_{p \in N(\nu_i)} \left(\omega_{_{n,{c_{j} \rightarrow c_{i} }}} {h}_{n,{c_{p} \rightarrow c_{i} }}^{\prime }\right) \\
\omega_{_{n,{c_{p} \rightarrow c_{i} }}} &= \text{MLP}\left({h}_{n,{c_{p} \rightarrow c_{i} }}^{\prime }\right)
\end{aligned}
\end{align}

Here, \( h_{n,\overrightarrow{ip}} \) represents the fusion of all relationship feature vectors associated with knowledge concepts \( i \). In the fusion process, each relationship feature vector \( {h}_{n,{c_{p} \rightarrow c_{i} }}^{\prime }\) associated with knowledge concepts \( i \) is first passed through a Multi-Layer Perceptron (MLP) to obtain the corresponding weight \( \omega_{_{n,{c_{p} \rightarrow c_{i} }}}\). Subsequently, the weight \( \omega_{_{n,{c_{p} \rightarrow c_{i} }}} \) is multiplied element-wise with the relationship feature vector \( {h}_{n,{c_{p} \rightarrow c_{i} }}^{\prime } \) and summed to produce the fused representation \( h_{n,\overrightarrow{ip}} \). Finally, the feature vector \( {h}_{n,{\overrightarrow{k}}} ^{\prime } \) of knowledge concepts \( i \) is concatenated with the fused representation \( h_{n,\overrightarrow{ip}} \), and after passing through the sigmoid activation function, the predecessor-successor cognitive structure representation \( \overrightarrow{h^{s}}\) for knowledge concepts \( i \) learned by learner \( n \) is generated.Based on this, the dependency cognitive structure representation of knowledge concepts \( i \) learned by learner \( n \) is defined as: \( \overleftrightarrow{h^{s}}\).

The final fused cognitive structure representation of the learner is defined as:
\begin{align}
    \begin{aligned}
{h^{s}} &= \text{sigmoid}\left({W}\left[\overleftrightarrow\omega\overleftrightarrow{h^{s}} \oplus\overrightarrow\omega\overrightarrow{h^{s}} \right] + b\right)  \\
\overleftrightarrow\omega &= \text{MLP}\left(\overleftrightarrow{h^{s}}\right)\\
\overrightarrow\omega &= \text{MLP}\left(\overrightarrow{h^{s}}\right)
\end{aligned}
\end{align}

where \(\overrightarrow{h^{s}}\) and \(\overleftrightarrow{h^{s}}\) are learned via functions that generate weights \(\overrightarrow{\omega}\) and \(\overleftrightarrow{\omega}\). The weights are then multiplied by the corresponding vectors, and the resulting values are concatenated. Finally, the cognitive structure representation \(h^{s}\) is derived via a sigmoid function.

\subsubsection{Prediction Module}

The output layer incorporates a classifier to predict learners' responses. The classifier maps input feature vectors to distinct categories, identifying patterns and rules from the training data to infer learners' response outcomes. The use of a classifier offers notable advantages: it leverages supervised learning techniques for efficient model training and effectively handles large-scale, high-dimensional feature spaces, thereby enhancing prediction accuracy.

In this study, a Multi-Layer Perceptron (MLP) is employed as the classifier for predicting learners' responses. The prediction task is formulated as a binary classification problem, where the response likelihood is mapped to a real number within the interval \([0, 1]\). If the predicted value is greater than or equal to 0.5, the learner is considered to have answered correctly; otherwise, the learner is considered to have answered incorrectly. This approach simplifies the problem representation while effectively utilizing feature space information to improve classification performance.

The designed MLP in this study features a three-layer hidden structure, with a sigmoid activation function applied at the end of each hidden layer. This design enables nonlinear feature mapping and facilitates information transfer between layers.

The fırst layer of the interaction layers is inspired by MIRT models. We formulate it as:
\begin{equation}
    x=Q_e\circ(h^s-h^{diff})\times h^{disc}
\end{equation}

where \(\circ\) is element-wise product. The following are two fully connected layers and an output layer:
\begin{align}
    \begin{aligned}
        x_{1} &= \phi(\mathbf{W}_1 \times \boldsymbol{x}^T + \boldsymbol{b}_1) \\
        x_{2} &= \phi(\mathbf{W}_2 \times \boldsymbol{x}_1 + \boldsymbol{b}_2) \\
        \hat{y} &= \phi(\mathbf{W}_3 \times \boldsymbol{x}_2 + b_3)
    \end{aligned}
\end{align}

where \(\phi\)is the activation function. Here we use Sigmoid.

In the model training phase, the parameters of the model are learned by minimizing the standard cross-entropy loss between the predicted probability of question correctness and the true label. Here, \( y \) represents the true answer outcome, with a value of 0 indicating an incorrect answer and a value of 1 indicating a correct answer.

\begin{equation}
    \ell = - \sum y \log(\hat{y}) + (1 - y) \log(1 - \hat{y})
\end{equation}

This loss function quantifies the difference between the predicted and actual outcomes, guiding the model in adjusting its parameters to improve prediction accuracy.
\section{Experiments}
In this section, we evaluate our CSCD framework on four popular and challenging real-world education datasets, comparing it to
the baselines. We also analyze the impact of interpretability.

\subsection{Dataset Description}
The dataset utilized in this study comprises three real-world datasets: ASSISTments2017, Junyi, and NIPS34.

The ASSISTments2017 dataset is a widely recognized public dataset in the field of online education, originating from the ASSISTments online learning platform. It primarily captures detailed behavioral data of learners during their response process. This dataset contains learners' response records, response times, knowledge concepts labels, and answer outcomes (correct or incorrect), with the goal of supporting personalized learning research and advancing cognitive diagnosis models.

The Junyi dataset is derived from an online learning platform targeting primary and secondary school learners, with an emphasis on tracking and analyzing learner learning behaviors and knowledge mastery. This dataset records various data related to learner practice, including knowledge concepts labels, response records (correct/incorrect), question types, and practice durations.

The NIPS34 dataset originates from Tasks 3 and 4 of the NeurIPS 2020 Education Challenge and includes learners' response records to multiple-choice diagnostic mathematics questions. The data were collected by the Eedi platform\cite{wang2020instructions}. For each question, the dataset identifies the leaf nodes from the subject tree as its corresponding knowledge components.

Detailed information regarding the datasets is provided in Table \ref{tableId1}.
\begin{table}[h]
\caption{Datasets summary.}%标题
\centering%把表居中
\begin{tabular}{cccc}%四个c代表该表一共四列，内容全部居中
\toprule%第一道横线
Datasets&ASSISTments2017&Junyi&NIPS34\\
\midrule%第二道横线 
\#learners&1,708&10,000&4,918 \\
\#Exercises&3,162&835&948 \\
\#Knowledges&102&835&57 \\
\#Response Logs&390,329&353,835&1,399,470 \\
\bottomrule%第三道横线
\end{tabular}
\label{tableId1}
\end{table}

\subsection{Experimental Settings}
\subsubsection{Baselines and Evaluation Metrics.}
To evaluate the effectiveness of our proposed RCD model, we compare it against several baseline methods. The details are presented as follows:
\begin{itemize}
    \item \textbf{IRT}\cite{lord1952theory}, being one of the most widely used CD methods, models the unidimensional features of learners and exercises using a linear function.
    \item \textbf{NCD}\cite{wang2020neural} is a recent deep learning-based CD model that models high-order, complex learner-exercise interaction functions using neural networks.
    \item \textbf{RCD}\cite{gao2021rcd} unifies the capture of inner structures and inter-layer interactions through a multi-layer relational graph, subsequently employing a multi-level attention network to integrate node-level relational aggregation within each local graph, while balancing graph-level relational aggregation across multiple graphs.
    \item \textbf{HierCD}\cite{li2022hiercdf} addresses the limitations of traditional attribute hierarchy models.
    \item \textbf{SCD}\cite{shen2024symbolic} incorporates symbolic trees to explicitly represent complex learner-exercise interaction functions, using gradient-based optimization methods to effectively learn both learner and exercise parameters.
\end{itemize}

To evaluate the performance of our framework, we utilize various metrics from both regression and classification perspectives. From the regression perspective, we use Root Mean Square Error (RMSE) to quantify the difference between predicted scores (i.e., continuous values ranging from 0 to 1) and actual values. From the classification perspective, we represent incorrect and correct learner answers as 0 and 1, respectively. Therefore, we use Prediction Accuracy (ACC) and Area Under the ROC Curve (AUC) to evaluate the framework.
\subsubsection{Parameter Settings.}
In our experiments, all learners are randomly divided into training, validation, and test sets in a ratio of 7:1:2. The batch size, dropout rate, and learning rate (with decay) are optimized over the ranges \{8, 16, 32, 64\}, (0, 0.5), and (1e-5, 2e-2), respectively. For fairness, the hyperparameters of the baseline models align with those reported in their respective papers and have been further fine-tuned to achieve optimal results. To configure the training process, we initialize the parameters using Xavier initialization and utilize flexible methods such as random search, grid search, and Bayesian search and selection strategies. If the AUC does not improve over the course of 10 epochs or if the maximum number of epochs (100) is reached, we implement an early stopping strategy. All experiments are conducted using PyTorch and executed on Linux servers equipped with RTX 4090 (24 GB).

\subsection{Overall Performance}
The experimental results, presented in the table \ref{tableId2}, demonstrate that \textbf{CSCD} outperforms all baseline models (e.g., IRT, NCD, RCD, HierCD, and SCD) across all three datasets (ASSISTments2017, Junyi, and NIPS34). Specifically, on the ASSISTments2017 dataset, \textbf{CSCD} achieves an AUC of 0.8010, an ACC of 0.7302, and an RMSE of 0.4248, surpassing all other models, particularly excelling in the AUC and ACC metrics. On the Junyi dataset, \textbf{CSCD} achieves an AUC of 0.8230, an ACC of 0.7667, and an RMSE of 0.3993, continuing to outperform other models, with notable advantages in RMSE, indicating superior predictive accuracy. On the NIPS34 dataset, \textbf{CSCD} achieves an AUC of 0.7889, an ACC of 0.7217, and an RMSE of 0.4257, maintaining strong performance, particularly in AUC and ACC, demonstrating its ability to effectively distinguish learner abilities and predict question correctness. Overall, \textbf{CSCD} consistently outperforms the baseline models across AUC, ACC, and RMSE metrics on all three datasets, highlighting its superior effectiveness, precision, and robustness in cognitive diagnosis tasks.

\begin{table}[h]
\caption{Experimental results on learner performance prediction.}
\centering
\begin{tabular}{cccccccccc}
\hline
   & \multicolumn{3}{c}{\textbf{ASSISTments2017}} & \multicolumn{3}{c}{\textbf{Junyi}} & \multicolumn{3}{c}{\textbf{NIPS34}} \\ \hline
   Model
        & AUC    & ACC    & RMSE    & AUC    & ACC    & RMSE    & AUC    & ACC    & RMSE    \\ \hline
IRT     & 0.7343 & 0.6773 & 0.4658  & 0.7409 & 0.7156 & 0.4342  & 0.7498 & 0.6970 & 0.4561  \\ 
NCD     & 0.7661 & 0.6992 & 0.4481  & 0.7786 & 0.7396 & 0.4229  & 0.7679 & 0.7013 & 0.4421  \\ 
RCD     & 0.7891 & 0.7186 & 0.4395  & 0.8095 & 0.7583 & 0.4175  & 0.7713 & 0.7156 & 0.4313  \\ 
HierCD  & 0.7761 & 0.7114 & 0.4348  & 0.7848 & 0.7439 & 0.4205  & 0.7756 & 0.7113 & 0.4613  \\ 
SCD     & 0.7543 & 0.6972 & 0.4566  & 0.7669 & 0.7486 & 0.4391  & 0.7583 & 0.6812 & 0.4593  \\ 
\textbf{CSCD} & \textbf{0.8010} & \textbf{0.7302} & \textbf{0.4248} & \textbf{0.8230} & \textbf{0.7667} & \textbf{0.3993} & \textbf{0.7889} & \textbf{0.7217} & \textbf{0.4257} \\ \hline
\end{tabular}
\label{tableId2}
\end{table}
\subsection{Ablation Study}
Ablation studies are commonly used control techniques in machine learning to validate the effectiveness of various components within a model. By progressively removing or disabling specific parts of the model and observing the resulting changes in performance, researchers can assess the contribution of each component to the model's overall performance. For example, if performance improves after adding modules A and B to the baseline model, but the performance with only module A is equivalent to or better than the performance with both modules A and B, it can be inferred that module B does not contribute additional improvements. Ablation studies provide valuable insights into the independence of research contributions and offer constructive feedback for model design. In deep learning, ablation studies typically involve removing certain components, such as network layers or features, to observe their impact on model performance, thereby helping researchers better understand the role and importance of each part of the model. This experiment utilizes ablation studies to evaluate the impact of the cognitive structure representation layer on the overall performance of the model.

The ablation study involves three datasets: ASSISTments2017, Junyi, and NIPS34. The experimental design includes different configurations of the cognitive structure representation layer, such as representations of only knowledge concepts, representations of only the relationships between knowledge concepts, and a comprehensive representation of the entire cognitive structure. These experiments aim to explore the influence of the overall cognitive structure representation on model performance. The experimental design is outlined in Table \ref{tableId3}.
\begin{table}[h]
\caption{Ablation Experiment Design.}%标题
\centering%把表居中
\begin{tabular}{ccc}%四个c代表该表一共四列，内容全部居中
\toprule%第一道横线
Name&knowledge concepts Representation&Relationship Representation\\
\midrule%第二道横线
CSCD w/s,g K& $\checkmark$ & $\times$ \\
CSCD w/s,g R& $\times$ & $\checkmark$ \\
CSCD w/s,g K+R& $\times$ & $\checkmark$ \\
\bottomrule%第三道横线
\end{tabular}
\label{tableId3}
\end{table}

In this context, "CSCD w/s,g K" indicates that the framework utilizes only the knowledge state within the cognitive structure state representation layer; "CSCD w/s,g R" signifies that the framework employs only the knowledge structure state within the same layer; and "CSCD w/s,g K+R" denotes that the framework integrates both knowledge state and the structure state within the cognitive structure representation layer, thereby utilizing the complete cognitive structure state representation. The experimental results are as follows:
\begin{figure}[h]
  \centering
  \includegraphics[width=1\linewidth]{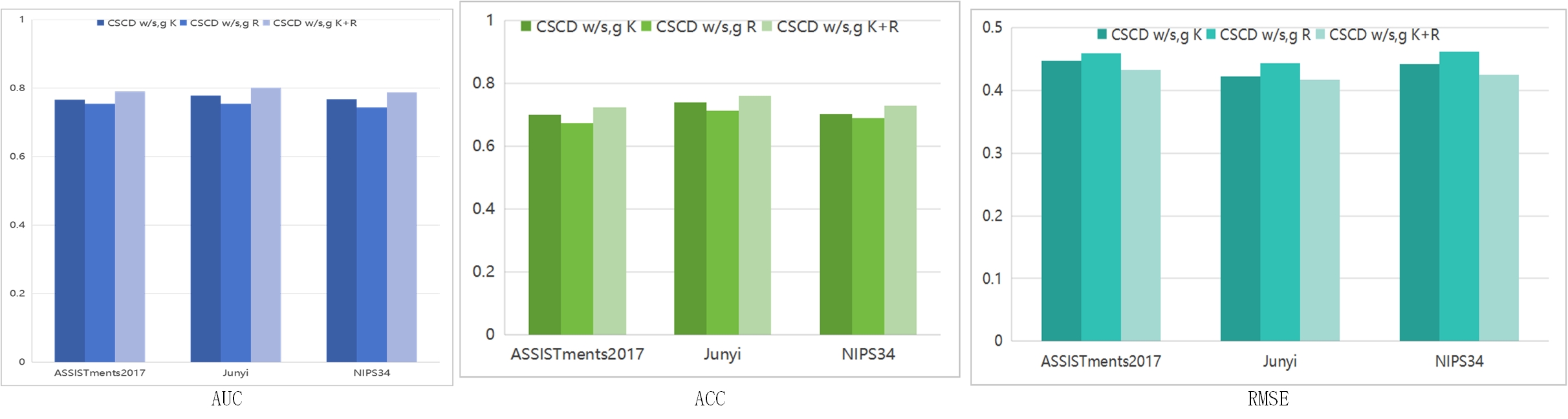}
  \caption{Ablation Experiment Results.}
  \label{exampleId4}
\end{figure}

The ablation experiment results, shown in Figure \ref{exampleId4} across three metrics, demonstrate that \textbf{CSCD w/s,g K+R} achieves the best performance on all three datasets (ASSISTments2017, Junyi, and NIPS34), thereby validating the crucial role of both the knowledge state and knowledge structure state in improving model performance. In terms of the AUC metric, \textbf{CSCD w/s,g K+R} outperforms both the model using only knowledge state representation (\textbf{CSCD w/s,g K}) and the model using only knowledge structure state representation (\textbf{CSCD w/s,g R}), indicating that the complete cognitive structure state representation more effectively captures learners' cognitive states and enhances the model’s discriminative ability. Regarding the ACC metric, \textbf{CSCD w/s,g K+R} also demonstrates a significant advantage, particularly on the Junyi and NIPS34 datasets, where performance improvements are most pronounced, further proving the effectiveness and stability of the complete cognitive structure state  representation. Additionally, in the RMSE metric, \textbf{CSCD w/s,g K+R} achieves the lowest prediction error, suggesting that the comprehensive representation significantly improves the model's predictive accuracy and reduces error accumulation. Overall, the ablation experiments clearly show that the complete cognitive structure state representation provides substantial advantages in cognitive diagnosis tasks.

\subsection{Interpretability and Visualization}
\begin{figure}[h]
  \centering
  \includegraphics[width=0.5\linewidth]{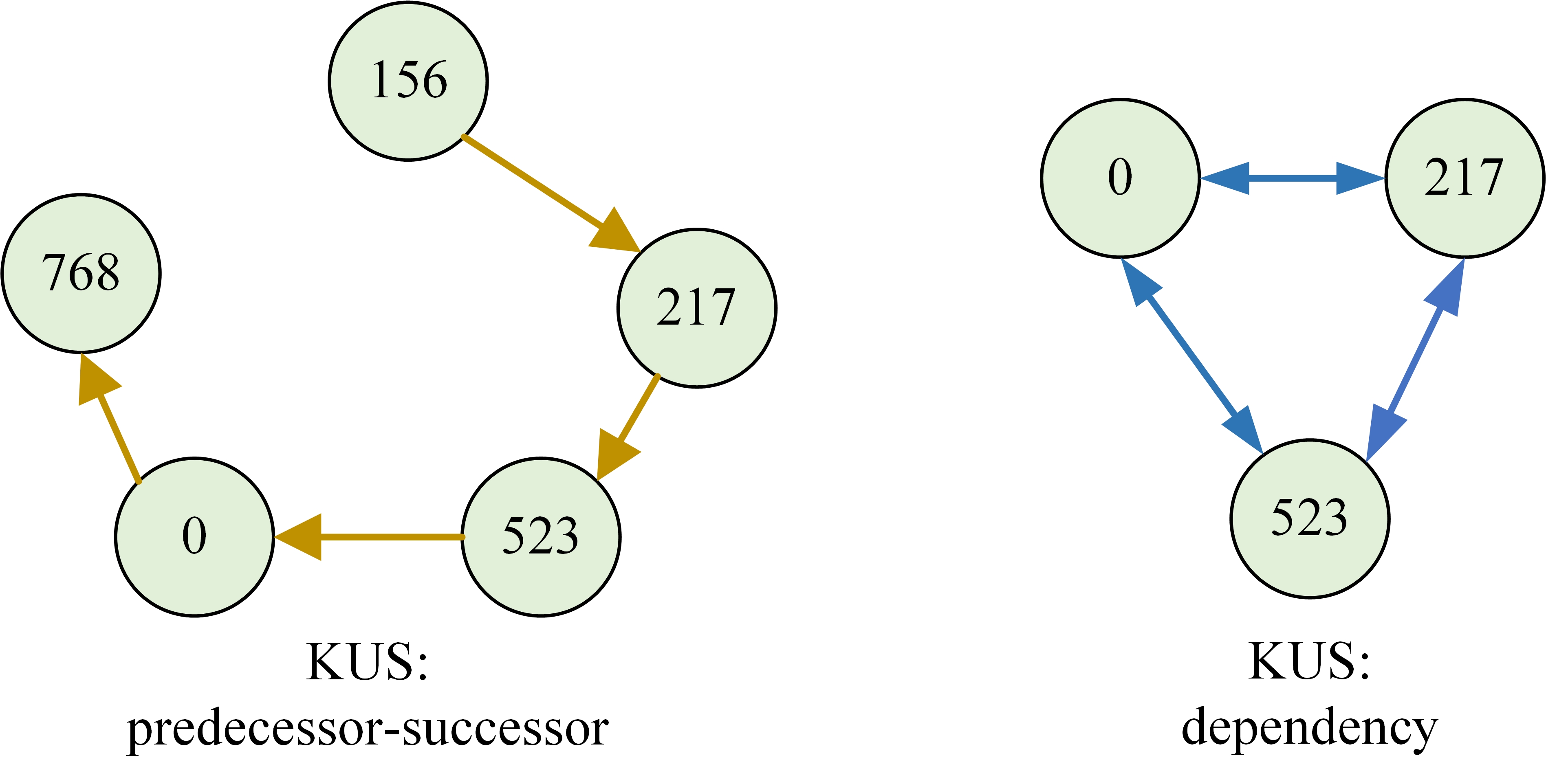}
  \caption{KUS: Predecessor-Successor and KUS: Dependency Relationships.}
  \label{exampleId5}
\end{figure}

\begin{figure}[h]
  \centering
  \includegraphics[width=1\linewidth]{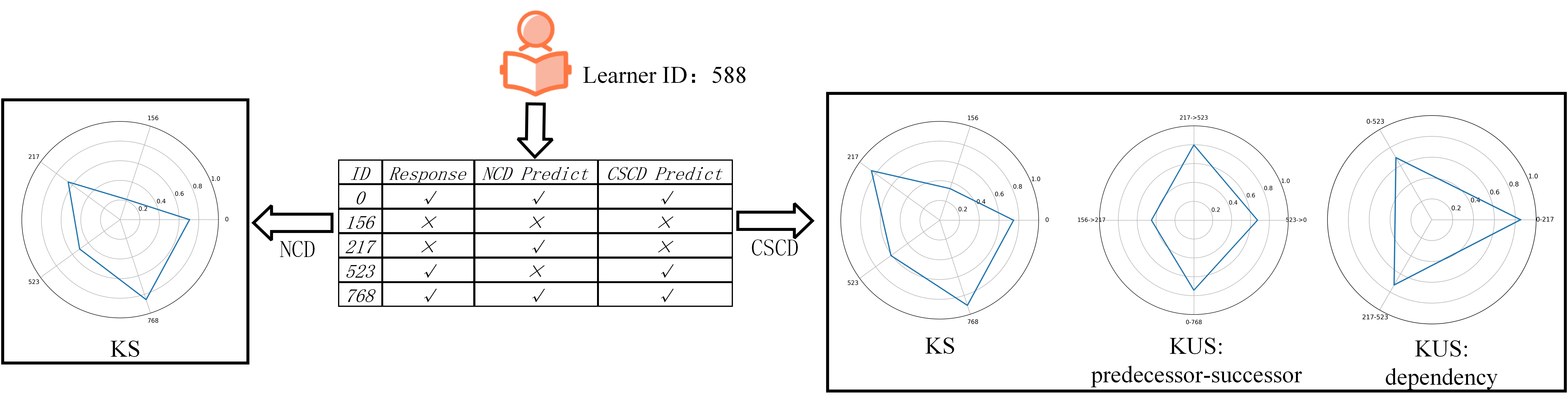}
  \caption{Example of a real interaction.}
  \label{exampleId6}
\end{figure}

We present an example of the \textbf{CSCD} diagnostic results for a learner in the Junyi dataset, shown in Figure \ref{exampleId6}, and compare it with the Neural Cognitive Diagnosis (NCD) model. The learner's cognitive level is diagnosed, and the corresponding radar chart is generated. Table \ref{tableId4} provides the knowledge concepts IDs and their corresponding names, while Figure \ref{exampleId5} illustrates the predecessor-successor and dependency relationships between the knowledge concepts.
\begin{table}[h]
\caption{knowledge concepts ID and name correspondence.}%标题
\centering%把表居中
\begin{tabular}{cc}%四个c代表该表一共四列，内容全部居中
\toprule%第一道横线
ID&Name\\
\midrule%第二道横线 
0&Parabola intuition \\
156&Equation of a line\\
217&Adding and subtracting polynomials\\
523&Recognizing conic sections \\
768&Parabola intuition \\
\bottomrule%第三道横线
\end{tabular}
\label{tableId4}
\end{table}

By analyzing the learner's response records, it was found that the learner made an error when answering knowledge concepts 217. The NCD model misjudged the learner's ability to correctly answer knowledge concepts 217, as evidenced by the radar chart, which indicates a high probability of the learner mastering this knowledge state(KS), leading to the misprediction. In contrast, the \textbf{CSCD} model accurately predicted the learner's performance across all exercises. Upon examining the radar chart generated by the \textbf{CSCD} model, which illustrates the learner's mastery of knowledge state(KS) and knowledge structure state(KUS), it becomes clear that, although the learner demonstrated a strong grasp of knowledge concepts 217, their understanding of the relationship between knowledge concepts 156 and 217 was weak. The \textbf{CSCD} model accounts for both the learner's mastery of individual knowledge state(KS) and the knowledge structure state(KUS), effectively modeling the complete cognitive structure, which led to more accurate predictions of the learner's performance. This demonstrates that incorporating knowledge concepts relationships into cognitive diagnosis models can significantly enhance prediction accuracy, further emphasizing the importance of the learner's cognitive structure in cognitive diagnosis tasks.

\section{Conclusion}
This paper addresses the limitations of existing cognitive diagnosis methods, which generally excel at diagnosing the knowledge state (KS) but fail to adequately model the learner's knowledge structure state (KUS). As a result, these methods do not provide a comprehensive representation of the learner's overall cognitive structure. To overcome this limitation, we propose a novel cognitive structure assessment framework based on an edge-feature-based graph attention network (EGAT). This framework dynamically models and updates both the knowledge state and knowledge structure state, thereby offering a holistic assessment of the learner's cognitive structure for diagnosing their performance on tasks.

The proposed model demonstrates significant improvements in both performance and interpretability, thereby supporting more effective personalized teaching strategies. The contributions of this paper include the introduction of the novel EGAT-based cognitive structure assessment framework and its validation through extensive experiments. We conducted multiple experiments on real-world educational datasets, comparing our model with existing methods, and demonstrating its superior performance in various diagnostic tasks.

%Bibliography
\bibliographystyle{unsrt}  
\bibliography{references}

\end{document}